\pdfoutput=1

\documentclass[11pt]{article}

\usepackage[]{ACL2023}

\usepackage{times}
\usepackage{latexsym}

\usepackage{tikz}
\usepackage[normalem]{ulem}
\usepackage{latexsym}
\usepackage{epsfig}
\usepackage{graphicx}
\usepackage[linesnumbered,ruled,vlined]{algorithm2e}
\usepackage{xcolor}

\usepackage{multirow}
\usepackage{etoolbox}
\usepackage{balance}
\usepackage{amsmath}
\usepackage{amsfonts}
\usepackage{balance}
\usepackage{pifont}
\usepackage{makecell}
\usepackage{hhline}
\usepackage{booktabs}
\usepackage{booktabs}
\usepackage{makecell}
\usepackage{graphicx}
\usepackage{mathtools}
\usepackage{enumitem}
\usepackage{float}
\usepackage{colortbl}
\usepackage[most]{tcolorbox}
\usepackage{pifont}
\usepackage{subcaption}
\usepackage{hyperref}
\usepackage{algorithmic}
\usepackage{cancel}

\usepackage[T1]{fontenc}

\usepackage[utf8]{inputenc}

\usepackage{microtype}

\usepackage{inconsolata}

\title{Counterfactual-Consistency Prompting for Relative Temporal Understanding in Large Language Models}

\newcommand\correspondingauthor{\thanks{~~Corresponding author.}}
\author{Jongho Kim and Seung-won Hwang\correspondingauthor \\
Interdisciplinary Program in Artificial Intelligence, Seoul National University \\ 
\texttt{\{jongh97, seungwonh\}@snu.ac.kr} \\
}

\newcommand{\ours}{CCP\xspace}

\newcommand{\contrast}{counterfactual\xspace}

\newtcolorbox[auto counter,number within=chapter,]{myColorbox}[3][]{
arc=5mm,
lower separated=false,
fonttitle=\bfseries,
colbacktitle=blue!10,
coltitle=blue!50!black,
enhanced,
attach boxed title to top left={xshift=0.5cm,
        yshift=-2mm},
colframe=blue!50!black,
colback=blue!10,
title=#2,#1,breakable}

\begin{document}

\maketitle
\begin{abstract}

Despite the advanced capabilities of large language models (LLMs), their temporal reasoning ability remains underdeveloped.
Prior works have highlighted this limitation, particularly in maintaining temporal consistency when understanding events. For example, models often confuse mutually exclusive temporal relations like ``before'' and ``after'' between events and make inconsistent predictions.
In this work, we tackle the issue of temporal inconsistency in LLMs by proposing a novel counterfactual prompting approach.
Our method generates counterfactual questions and enforces collective constraints, enhancing the model's consistency.
We evaluate our method on multiple datasets, demonstrating significant improvements in event ordering for explicit and implicit events and temporal commonsense understanding by effectively addressing temporal inconsistencies.

\end{abstract}
\section{Introduction}

Despite the impressive capabilities of LLMs,
a line of research~\cite{jain-etal-2023-language-models,chu2023timebench} has highlighted that these models often lack temporal reasoning abilities. This is especially true for \textit{relative} event understanding, where the goal is to infer temporal relationships between events or properties within an event in the passage, without depending on \textit{absolute} time indicators (e.g., specific dates).

\begin{figure}[t]
    \centering
    \includegraphics[width=1\linewidth]{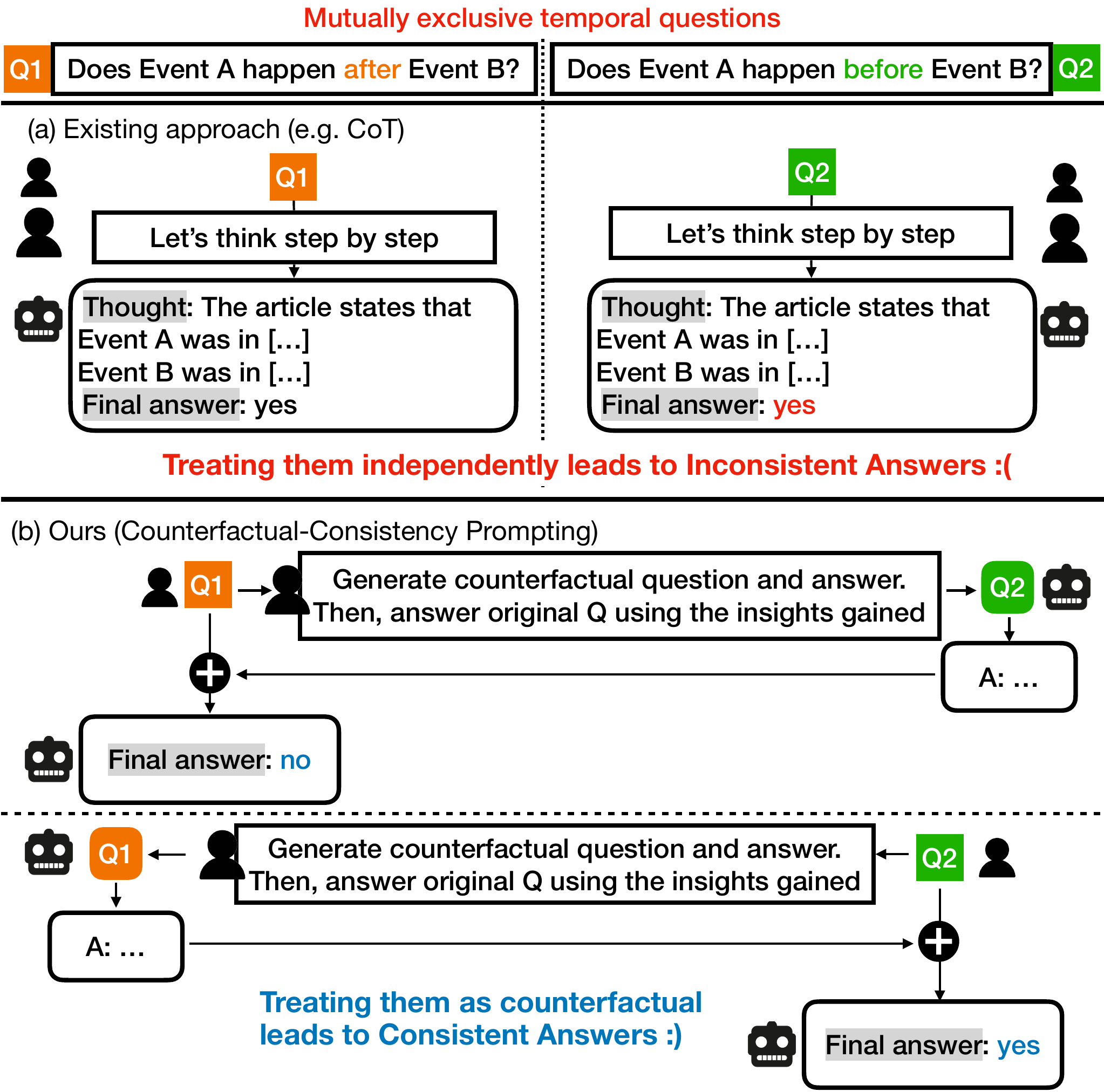}
    \caption{Example of leveraging \contrast questions to resolve temporal inconsistency in LLMs.}
    \label{fig:ours}
\end{figure}

The primary challenge is that LLMs lack \textit{temporal consistency} in their responses~\cite{qiu-etal-2024-large, chen2024improving}. Temporal consistency is defined as the model's ability to ensure that conflicting timelines do not co-exist.
For instance in Figure~\ref{fig:ours}-(a), if the model is temporally inconsistent, mutually exclusive temporal relations like ``before'' and ``after'' are sometimes confused when ordering events, leading to contradictory predictions—such as stating that Event A happens both before and after Event B in the same context.

While events with time indicators are often addressed with mathematical reasoning~\cite{zhu2023question,su2024timo}, no existing work has successfully tackled the challenge of temporal inconsistency in the events' relative understanding without requiring explicit time markers. 
Chain-of-thought (CoT) reasoning~\cite{wei2022chain}, which primarily aids mathematical and symbolic reasoning~\cite{sprague2024cotmainlymath}, is also reported 
to fail to solve such inconsistency~\cite{qiu-etal-2024-large}.
Considering temporal consistency is fundamental in temporal reasoning, its absence in LLM can undermine key tasks like planning~\cite{sakaguchi-etal-2021-proscript-partially,zhang-etal-2024-narrative}.
These observations highlight the need for alternative reasoning skills to achieve temporal consistency.

This study answers the following research question: \textbf{Can we prompt LLMs to elicit the ability to mitigate temporal inconsistency?} 
Inspired by counterfactual augmentation,
where models are exposed with lexically similar, but typically label-flipping pairs in training
\cite{kaushik2020learning},
we extend it to LLMs to generate \textit{temporally \contrast questions}: 
We introduce lexically small interventions to the original input (e.g. before to after, years to centuries) that drastically affect its temporal semantics. By providing these questions and self-generated answers alongside the original input, the model would rely less on lexical similarities and better understand the semantics.

To this end, we propose a novel counterfactual-consistency prompting (\ours), designed to enhance the temporal consistency of LLMs, as described in Figure~\ref{fig:ours}-(b). \ours first generates temporal counterfactual exemplars and then applies the insights gained to address the original temporal question.
This method is particularly effective in relative event understanding because the counterfactual exemplars not only encourage the model to understand different temporal semantics but also directly impose temporal constraints.
For instance, if the model states that ``Event A happens after Event B'' and also recognizes that ``Event A happens before Event B'', the conflict forces the model to collectively re-weight the validity of these two statements.

We show performance gain of \ours across multiple relative event understanding tasks. Our effectiveness in mitigating temporal inconsistencies is further demonstrated by our inconsistency metric.

\begin{table*}[h]
\centering
\scalebox{0.85}{
\begin{tabular}{|l|l|l|l|}
\hline
\textbf{Dataset} & \textbf{Relation} & \textbf{Revision Type} & \textbf{Example} \\ \hline
TempEvalBi-QA & $r_1(e_1,e_2)$ & Original & Did they get married \textbf{after} they moved to Maine? \\ 
 & $r_2(e_1,e_2)$ & $r_1 \rightarrow r_2$ & Did they get married \textbf{before} they moved to Maine? \\ \hline
TRACIE & $r_1(e_1,e_2)$ & Original & Chad found the cap \textbf{before} he got off of the ride. \\ 
 & $r_2(e_1,e_2)$ & $r_1 \rightarrow r_2$ & Chad found the cap \textbf{after} he got off of the ride. \\ \hline
MCTACO: Duration & $r_1(e_1)$ & Original & It has existed for \textbf{1 year}. \\ 
 & $r_2(e_1)$ & $r_1 \rightarrow r_2$ & It has existed for \textbf{centuries}. \\ \hline
MCTACO: Frequency & $r_1(e_1)$ & Original & It rained \textbf{a couple of times} in the summer. \\ 
 & $r_2(e_1)$ & $r_1 \rightarrow r_2$ & It rained \textbf{every month} in the summer. \\ \hline
MCTACO: Stationarity & $r_1(e_1)$ & Original & She is \textbf{still} in Ranchipur. \\  
 & $r_2(e_1)$ & $r_1 \rightarrow r_2$ & She is \textbf{not} in Ranchipur. \\ \hline
MCTACO: Ordering & $r_1(e_1,e_2)$ & Original & They \textbf{went to the store} after they were put in jail. \\  
 & $r_1(e_1,e_3)$ & $e_2 \rightarrow e_3$ & They \textbf{repented} after they were put in jail. \\ \hline
MCTACO: Typical Time & $r_1(e_1)$ & Original & Durer died \textbf{40 years} later. \\  
 & $r_2(e_1)$ & $r_1 \rightarrow r_2$ & Durer died \textbf{360 years} later. \\ \hline
\end{tabular}
}
\vspace{-3mm}
\caption{The temporal aspects targeted by each dataset and types of counterfactuals addressed by our framework.
The examples illustrate how counterfactual questions modify the semantics regarding temporal relations ($r_1$,$r_2$) for events ($e_1$,$e_2$,$e_3$).
}
\vspace{-5mm}
\label{tab:full_contrast_examples}
\end{table*}

\section{Method}

Our idea is to make the model approximate the \textit{temporal constraints} using counterfactuals. 
Because temporal reasoning imposes unique interdependencies, where one temporal aspect affects another~\cite{han2019deep,kim-etal-2024-chaining}, counterfactuals enable us to capture these constraints.
For example, if the model establishes from a counterfactual exemplar that ``Event $e_1$ happens [$r_2$: before] Event $e_2$'', it is constrained to predict the original question that ``Event $e_1$ cannot happen [$r_1$: after] Event $e_2$'':
\begin{equation}
r_2(e_1, e_2) \in \mathcal{V} \implies r_1(e_1, e_2) \not\in \mathcal{V}
\end{equation}
where $r(e_a,e_b)$ represents the temporal relation $r$ between events $e_a$ and $e_b$, and $\mathcal{V}$ represents the set of coherent temporal relations with the context.

\subsection{Generating Temporally Counterfactual Questions}

Given a context $C$, our task is to provide an accurate answer to question $Q$ while maintaining temporal consistency.
We start by creating \textit{temporally counterfactual} questions, $Q^{c_1}...Q^{c_i}$.

Counterfactual augmentation conventionally aims to generate instances with lexically minimal edits
while keeping others unchanged~\cite{huang2019reducing,kaushik2020learning,wang2020identifying}. This strategy discourages models from relying too much on superficial similarity. However, previous works~\cite{kaushik2020learning} require arbitrary label-flipping edits, which are unsuitable for LLM inference with unknown test labels.

In contrast, we focus on \textit{temporally counterfactual} questions that specifically edit the temporal semantics of the original sentence.  
Our approach ensures that the model adheres to the ``temporal constraints'', yet retains the effect of label-flipping that emphasizes temporal cues over superficial similarity.

Specifically, we set the types of counterfactual questions based on the temporal semantics each dataset aims to capture.
The types of temporal counterfactuals are listed in Table~\ref{tab:full_contrast_examples}. 
For sentences representing temporal relation between two events ($r_1(e_1,e_2)$), we substitute the relation $r_1$ to $r_2$, or event $e_2$ to $e_3$. For those representing an event's temporal properties ($r_1(e_1)$) such as duration or stationarity, we substitute the property to $r_2$ or negate it to $\neg r_1$.

To generate counterfactual questions, we design our model to dynamically create them rather than rely on a predefined rule-based template. While rule-based approaches like \citet{chen2024improving} in logical reasoning constrain answers using a predefined question set, they limit flexibility to cover the broad range of temporal expressions. By comparison, our dynamic generation of constraints provides a more adaptable solution.
We specify various counterfactual types through in-context learning (ICL) to control the relevance of these generated questions. The full prompts are in~Appendix~\ref{appndx:prompt_examples_gen_counter_q}.

\subsection{Counterfactual-Consistency Prompting}

After generating the counterfactual questions, we prompt the model again to produce predictions for counterfactual $Y^{c_1},...,Y^{c_{n}}$.
However, there is a risk when LLMs may fail to answer the counterfactual questions correctly.
In this case, their direct use propagates errors to the original question. 

As a proxy for determining whether the generated prediction can be trusted, existing works aggregate multiple predictions of the same question~\cite{wang2023self,du2024improving}.
Formally, the refined prediction $Y$ is derived by re-weighting the probability distribution $P$ of previous predictions $Y_{1},...,Y_{n}$ from the same question as:
$P(Y) = f(P(Y_{1}), ..., P(Y_{n}))$
where $f$ is an aggregation function such as majority voting or LLM itself. Though, they can lead to errors as they solely rely on feedback from a single question.

Our distinction is to aggregate predictions from both the original and counterfactual questions. We design the model to re-weight the counterfactual answer distributions across the questions. 
\begin{equation}
\scalebox{0.85}{$P(Y) = f(P(Q,Y), P(Q^{c_1}, Y^{c_1}), ..., P(Q^{c_n}, Y^{c_n}))$}
\end{equation}
For instance, even if the model wrongly predicts the relation as `after' in a counterfactual, collectively considering the possibility of the relation `before' can re-weight the effect of the constraint. The prompts are provided in Appendix~\ref{appndx:prompt_examples_gen_ans}.

This re-evaluation approach improves robustness against potential errors in generated answers.
The second analysis in Subsection~\ref{subsec:anal} shows such self-correction outperforms a baseline directly leveraging counterfactuals without aggregation.

\section{Experiments}

\begin{table*}[ht]
\centering
\scalebox{1.0}{
\begin{tabular}{c|c|ccc|ccc|ccc}
\toprule
 &            & \multicolumn{3}{c|}{TempevalQA-Bi}                                         & \multicolumn{3}{c|}{TRACIE}                                               & \multicolumn{3}{c}{MCTACO}    \\ \midrule
 &            & ACC                     & F1                     & INC ($\downarrow$)      & ACC                     & F1                      & INC ($\downarrow$)     &   ACC                & F1                   &     INC ($\downarrow$)        \\ \midrule
Llama   & SP         & 65.4                    & 63                      & 57.6                   & 57.4                    & 66.9                   & 75.2                    & 77.7                 & 69.4                 & 59.8                 \\
-3-8B & CoT        & 69.6                    & 70.6                    & 50                     & 63                      & 64.9                   & 56                      & 77.6                 & 69.8                 & 63.4                 \\
        & Consistency    & 70.8                    & 71.2                    & 49.6                   & 64.9                    & 67.3                   & 57.8                    & 77.5                 & 69.0                 & 61.1                 \\
        & Reflection & 63.6                    & 63.9                    & 44.6                   & 62.5                    & 55.7                   & 55.5                    & 77.4                 & 69.7                 & 76.4                 \\
        & Debate     & 67.6                    & 65.2                    & 52.2                   & 63.6                    & 66                     & 53.2                    & 37.4                 & 31.6                 & 88.1                 \\
        & CCP        & \textbf{75.9}          & \textbf{75.2}           & \textbf{32.7}      & \textbf{68.8}      & \textbf{70.4}      & \textbf{39.8}      & \textbf{78.7}                 & \textbf{81.7}                 & \textbf{57.7}                   \\ \midrule
Llama   & SP         & 76.6 & 78.6 & 39.7 & 79.9 & 79.7 & 29.6 & 85.2 & 81.8 & 43.5 \\
 -3-70B & CoT        & 80.4 & 82   & 31.3 & 80.1 & 80   & 31.8 & 85.9 & 82.2 & 46.9 \\
 & Consistency   & -    & -    & -    & -    & -    & -    & -    & -    & -    \\
 & Reflection & 77   & 77.9 & 35.3 & 80   & 78.3 & 30.3 & 80.6 & 73   & 56.5 \\
 & Debate     & 81   & 82.8 & 32.6 & 81.6 & 80.7 & 25.9 & 85.3 & 81.4 & 45.9 \\
 & CCP        & \textbf{87.3} & \textbf{87.9} & \textbf{19.2} & \textbf{86.5} & \textbf{86.1} & \textbf{12.0}   & \textbf{89.4} & \textbf{87.6} & \textbf{37.0} \\ \midrule
GPT-4o & SP         & 78.8                    & 76.4                    & 36.6                   & 74.6                    & 71.3                   & 38.2                    & 76.0                   & 63.1                 & 65.8                 \\
-mini   & CoT        & 81.3                    & 79.9                    & 29                     & 73.2                    & 68.5                   & 42.7                    & 80.9                 & 73.7                 & 58.9                 \\
        & Consistency& 85.5                    & 85.5                    & 21.9                   & 73.6                    & 68.8                   & 42.8                    & 78.9	               & 69.4                 & 60.6             \\
        & Reflection & 86.8                    & 86.9                    & 22.8                   & 74.4                    & 70.9                   & 39.1                    & 74.8                 & 60.2                 & 68.5                 \\
        & Debate     & 86.4                    & 86.4                    & 24.6                   & 73                      & 67.1                   & 44.5                    & 78.3                 & 68.2                 & 61.0                   \\
       & CCP        & \textbf{88.8}            & \textbf{88.7}          & \textbf{19.6}           & \textbf{82.5}           & \textbf{81.2}          & \textbf{20.2}           & \textbf{87.8}       & \textbf{85.8}         & \textbf{42.1}        \\ \midrule
GPT-4o & SP         & 86.4                    & 85.8                    & 20.1                   & 80.1                    & 78.6                   & 27.0                      & 79.7                 & 70.9                 & 60.5                 \\
       & CoT        & 90.4                    & 90                      & 17.4                   & 80.2                    & 78.1                   & 32.4                    & 84.4                 & 80                   & 49.7                 \\
       & Consistency& 91.7                    & 91.5                    & 14.7                   & 80.1                    & 77.7                   & 31.4                    & 82.9                 & 77.3                 & 49.7                 \\
       & Reflection & 93.1                    & 93                      & 11.2                   & 82.7                    & 80.9                   & 26.6                    & 80.0                   & 72.2                 & 55.4                 \\
       & Debate     & 90.8                    & 90.6                    & 11.2                   & 80.6                    & 77.9                   & 32.8                    & 81.4                 & 74.6                 & 52.2                 \\
       & CCP        & \textbf{93.8}           & \textbf{93.8}          & \textbf{8.0}          & \textbf{85.8}           & \textbf{84.7}         & \textbf{17.6}              & \textbf{90.4}             & \textbf{88.8}          & \textbf{35.0}                 \\ \bottomrule
\end{tabular}
}
\caption{The full performance comparison results on the relative event understanding tasks. Our prompting methods, which leverage self-generated exemplars as the temporal constraint, outperform baselines across the board.}
\label{tab:main}
\vspace{-5mm}
\end{table*}

\subsection{Datasets}

Among publicly available datasets, we selected three based on two criteria: (1) the task focuses on relative event understanding without absolute time indicators, and (2) the temporal inconsistency on the dataset can be evaluated.

\textbf{TempEvalQA-Bi}~\cite{qiu-etal-2024-large} involves ordering two explicit events in time, assessing temporal consistency in mutually exclusive question pairs.
\textbf{TRACIE}~\cite{zhou2021temporal} expands the event ordering to implicit events, testing if the hypothesis logically follows the story. 
We finally added \textbf{MCTACO}~\cite{zhou2019going} considering the diverse event-related temporal properties. 
The dataset covers broader aspects like event duration or frequency.
We modified the multiple-choice setting of MCTACO into a binary question-answering task for consistency
evaluation, presenting each answer candidate separately to determine if it fits the context. Dataset statistics and examples are in Appendix~\ref{appndx:data_summary}.

\subsection{Metrics}
Along with accuracy (ACC) and F1 scores to assess overall performance, we introduce the inconsistency metric (INC) as a main evaluation measure for temporal inconsistency. We define the INC as the percentage of inconsistent predictions. An inconsistency is counted when at least one incorrect answer is found within a group of minimally dissimilar questions with slight modifications in their temporal semantics, while all other aspects remain unchanged.

TempEvalQA-Bi directly provides this metric. For TRACIE, we manually group questions that are counterfactual to each other. 
We adapt INC in MCTACO by grouping original multiple-choice candidates by question. 

\subsection{Evaluation Settings and Baselines}

For models, we used open-source LLM \textbf{Llama-3 8B} and \textbf{70B}~\cite{llama3modelcard}, and API-based LLM \textbf{GPT-4o-mini} and \textbf{GPT-4o}~\cite{openai2024gpt4technicalreport}. 

For baselines, we first compare \ours with standard prompting (\textbf{SP}) that directly answers the question without intermediate steps, and \textbf{CoT}, which incorporates step-by-step reasoning to derive the answer. Next, we consider methods that aggregate multiple predictions of the same question. Self-\textbf{Consistency}~\cite{wang2023self} predicts one question multiple times and performs majority voting. Self-\textbf{Reflect} methods~\cite{madaan2024self,shinn2024reflexion} iteratively refine own predictions. Multi-agent \textbf{Debate}~\cite{du2024improving} leverages both majority vote and reflection. We employ a 3-shot setting across all configurations. More details on evaluation settings are in Appendix~\ref{appndx:hyperparam}.

\subsection{Main results}

Table~\ref{tab:main} highlights the performance of our method compared to baseline methods on relative event understanding tasks.
Compared to SP, the CoT baseline is not usually effective and often worsens performance. 
Advanced baselines, Consistency, Reflect, and Debate, also fail to consistently reduce inconsistencies or achieve competitive accuracy.
In contrast, \ours steadily outperforms these baselines across various LLMs, significantly reducing temporal inconsistencies across all datasets and achieving notable improvements in ACC and F1 scores.

\subsection{Analysis}
\label{subsec:anal}
\paragraph{Creating \contrast questions by generation handles diverse temporal aspects.} We compared our generative setting with the \textbf{Ret}rieved \textbf{Q}uestions (Ret.Q) approach, where \contrast questions were retrieved from other questions within the same question group.
We evaluated the methods on MCTACO, which covers various aspects of event reasoning.

\begin{figure}[h]
    \centering
    \includegraphics[width=1\linewidth]{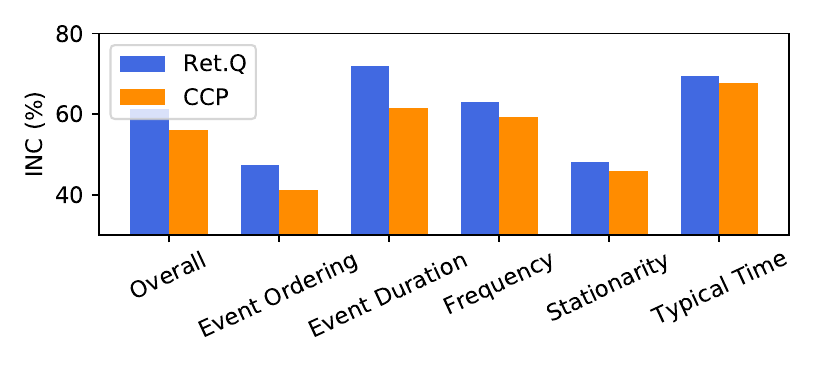}
    \caption{Comparison between counterfactual example collection methods on MCTACO with Llama-3-8B.}
    \label{fig:mctaco_each_type}
\end{figure}

Figure~\ref{fig:mctaco_each_type} shows that generating \contrast questions proved more effective for all temporal types.
These results suggest that our method performs better in event understanding with diverse relations, where the dataset cannot often provide high-quality counterfactual questions. Notably, \ours outperforms the Ret.Q baseline even though our method may produce incorrect questions. Also, \ours is more practical since Ret.Q assumes the questions in the test set are observed.

\begin{figure}[h]
    \centering
    \includegraphics[width=1\linewidth]{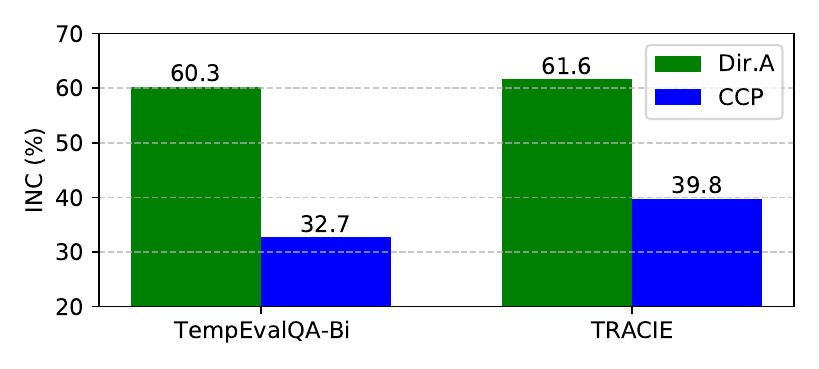}
    \caption{Comparison between different counterfactual leveraging methods with the Llama-3-8B model.}
    \label{fig:ours_vs_oursd}
\end{figure}

\paragraph{CCP is robust against wrong counterfactual exemplars.}
We conducted a comparative analysis of two methods: \textbf{Dir}ect \textbf{A}nswering (Dir.A), which involves answering directly from counterfactual exemplars, versus $\text{\ours}$ which leverages the aggregation step to re-evaluate them. 
We conducted experiments on TempEvalQA-Bi and TRACIE, where before-after relations ensure that identifying a counterfactual answer is sufficient to determine the original. We excluded MCTACO since its counterfactual answers do not always determine the validity of the original answer. In the $\text{Dir.A}$ implementation, the answer to the counterfactual question is flipped and directly used as the response to the original question.
The results in Figure~\ref{fig:ours_vs_oursd} demonstrate that \ours consistently outperforms $\text{Dir.A}$, supporting our robustness by the collective evaluation.

\paragraph{CCP's performance gain is orthogonal to the number of ICL examples.}

Our approach inevitably introduces additional counterfactual examples during in-context learning (ICL), leading to a higher total number of shots compared to the baseline. To ensure a more competitive baseline, we increased the total number of shots in the baseline. In the MCTACO dataset and with the Llama model, we additionally experimented with the 12-shot CoT, which includes 12 passage (P)-question (Q)-candidate (C) pairs, and compared them with our 3-shot. We note that our 3-shot examples include 3 passage-question pairs and 11 candidates.

The results in Table~\ref{tab:moreshot} demonstrate that our method significantly outperforms the CoT, even with the increased number of examples in the baseline (INC score: 60.0 for CoT vs. 57.7 for Ours). This indicates that the performance gains are not simply due to the inclusion of more examples but are primarily driven by leveraging temporal constraints through counterfactual questions to enhance reasoning.

Additionally, we tested whether our approach benefits from additional ICL examples. The results in the last row of Table~\ref{tab:moreshot} confirm this, showing an improvement in INC score from 57.7 to 49.8, further validating the potential performance gain of our method.
\begin{table}[h]
\centering
\scalebox{1.0}{
\begin{tabular}{c|cc|ccc}
\toprule
             & \multicolumn{5}{c}{MCTACO}    \\ \midrule
       & \#P-Q  & \#C     &  ACC          & F1           & INC ($\downarrow$)            \\ \midrule
 CoT  &  3     & 3           & 77.6        & 69.8      & 63.4          \\
 CoT  &  12     & 12         & 78.9        &   72.2    &   60.0      \\ \midrule
 \ours &  3    & 11          & 78.7  & 81.7 & 57.7 \\ 
 \ours &   12   & 26    & \textbf{85.0} & \textbf{82.2} & \textbf{49.8} \\
 \bottomrule
\end{tabular}
}
\caption{Performance comparison of Llama-3-8B on MCTACO with the different number of ICL examples.}
\label{tab:moreshot}
\end{table}
\section{Conclusion}

We targeted the temporal inconsistency in relative event understanding with LLMs by proposing a prompting approach using \contrast questions.
This encourages the model to focus more on the temporal aspects and collectively evaluate its answer with imposed constraints.
Experiments with the INC metric show that our approach mitigates inconsistency and improves overall performance.

\section{Limitation}

Our method showed limited performance improvement when time indicators, such as specific years (e.g., 1980), are involved in temporal understanding. This is implied from our evaluations on event-time ordering and time-time ordering tasks, as shown in Appendix~\ref{appndx:detail_limitation_temporal_indicator}. The findings suggest that arithmetic reasoning is essential for grounding timelines with absolute time indicators, as emphasized in prior studies~\cite{su2024timo,zhu2023question}. 

Another limitation is that we focused on pointwise and pairwise event reasoning to highlight the model's struggles with basic temporal reasoning due to consistency issues. We anticipate future work expanding our approach to more complex listwise ordering like event schema prediction~\cite{zhang-etal-2024-narrative}.

Finally, the effectiveness of LLMs can be impacted by the specific wording of prompts~\cite{jiang2020can}. For example, Appendix~\ref{appndx:detail_limitation_prompt_sensitivity} shows that, although our revised prompts generally yield better results than CoT, \ours experienced a slight decrease in performance.

\section*{Acknowledgement} 
This work was supported by Naver, the National Research Foundation of Korea(NRF) grant funded by the Korea government(MSIT) (No. RS-2024-00414981), Institute of Information \& communications Technology Planning \& Evaluation (IITP) grant funded by the Korea government (MSIT) (No. 2022-0-00077/RS-2022-II220077, AI Technology Development for Commonsense Extraction, Reasoning, and Inference from Heterogeneous Data), and Institute of Information \& communications Technology Planning \& Evaluation (IITP) grant funded by the Korea government(MSIT) [NO.RS-2021-II211343, Artificial Intelligence Graduate School Program (Seoul National University)].
\bibliography{anthology,custom}

\begin{thebibliography}{31}
\expandafter\ifx\csname natexlab\endcsname\relax\def\natexlab#1{#1}\fi

\bibitem[{AI@Meta(2024)}]{llama3modelcard}
AI@Meta. 2024.
\newblock \href {https://github.com/meta-llama/llama3/blob/main/MODEL_CARD.md} {Llama 3 model card}.

\bibitem[{Chen et~al.(2024)Chen, Ma, Song, Cao, Zhang, and Li}]{chen2024improving}
Meiqi Chen, Yubo Ma, Kaitao Song, Yixin Cao, Yan Zhang, and Dongsheng Li. 2024.
\newblock Improving large language models in event relation logical prediction.
\newblock In \emph{Proceedings of the 62nd Annual Meeting of the Association for Computational Linguistics (Volume 1: Long Papers)}, pages 9451--9478.

\bibitem[{Chen et~al.(2021)Chen, Wang, and Wang}]{chen2dataset}
Wenhu Chen, Xinyi Wang, and William~Yang Wang. 2021.
\newblock A dataset for answering time-sensitive questions.
\newblock In \emph{Thirty-fifth Conference on Neural Information Processing Systems Datasets and Benchmarks Track (Round 2)}.

\bibitem[{Chu et~al.(2023)Chu, Chen, Chen, Yu, Wang, Liu, and Qin}]{chu2023timebench}
Zheng Chu, Jingchang Chen, Qianglong Chen, Weijiang Yu, Haotian Wang, Ming Liu, and Bing Qin. 2023.
\newblock Timebench: A comprehensive evaluation of temporal reasoning abilities in large language models.
\newblock \emph{arXiv preprint arXiv:2311.17667}.

\bibitem[{Du et~al.(2024)Du, Li, Torralba, Tenenbaum, and Mordatch}]{du2024improving}
Yilun Du, Shuang Li, Antonio Torralba, Joshua~B Tenenbaum, and Igor Mordatch. 2024.
\newblock Improving factuality and reasoning in language models through multiagent debate.
\newblock In \emph{Forty-first International Conference on Machine Learning}.

\bibitem[{Fang et~al.(2024)Fang, Li, Wang, Lin, and Feng}]{fang2024counterfactualdebatingpresetstances}
Yi~Fang, Moxin Li, Wenjie Wang, Hui Lin, and Fuli Feng. 2024.
\newblock \href {http://arxiv.org/abs/2406.11514} {Counterfactual debating with preset stances for hallucination elimination of llms}.

\bibitem[{Han et~al.(2019)Han, Hsu, Yang, Galstyan, Weischedel, and Peng}]{han2019deep}
Rujun Han, I-Hung Hsu, Mu~Yang, Aram Galstyan, Ralph Weischedel, and Nanyun Peng. 2019.
\newblock Deep structured neural network for event temporal relation extraction.
\newblock In \emph{Proceedings of the 23rd Conference on Computational Natural Language Learning (CoNLL)}, pages 666--106.

\bibitem[{Huang et~al.(2019)Huang, Zhang, Jiang, Stanforth, Welbl, Rae, Maini, Yogatama, and Kohli}]{huang2019reducing}
Po-Sen Huang, Huan Zhang, Ray Jiang, Robert Stanforth, Johannes Welbl, Jack Rae, Vishal Maini, Dani Yogatama, and Pushmeet Kohli. 2019.
\newblock Reducing sentiment bias in language models via counterfactual evaluation.
\newblock \emph{arXiv preprint arXiv:1911.03064}.

\bibitem[{Jain et~al.(2023)Jain, Sojitra, Acharya, Saha, Jatowt, and Dandapat}]{jain-etal-2023-language-models}
Raghav Jain, Daivik Sojitra, Arkadeep Acharya, Sriparna Saha, Adam Jatowt, and Sandipan Dandapat. 2023.
\newblock \href {https://doi.org/10.18653/v1/2023.emnlp-main.418} {Do language models have a common sense regarding time? revisiting temporal commonsense reasoning in the era of large language models}.
\newblock In \emph{Proceedings of the 2023 Conference on Empirical Methods in Natural Language Processing}, pages 6750--6774, Singapore. Association for Computational Linguistics.

\bibitem[{Jiang et~al.(2020)Jiang, Xu, Araki, and Neubig}]{jiang2020can}
Zhengbao Jiang, Frank~F Xu, Jun Araki, and Graham Neubig. 2020.
\newblock How can we know what language models know?
\newblock \emph{Transactions of the Association for Computational Linguistics}, 8:423--438.

\bibitem[{Kaushik et~al.(2020)Kaushik, Hovy, and Lipton}]{kaushik2020learning}
Divyansh Kaushik, Eduard Hovy, and Zachary Lipton. 2020.
\newblock Learning the difference that makes a difference with counterfactually-augmented data.
\newblock In \emph{International Conference on Learning Representations}.

\bibitem[{Kim et~al.(2024)Kim, Lee, Kim, and Hwang}]{kim-etal-2024-chaining}
Jongho Kim, Dohyeon Lee, Minsoo Kim, and Seung-won Hwang. 2024.
\newblock \href {https://aclanthology.org/2024.eacl-long.101} {Chaining event spans for temporal relation grounding}.
\newblock In \emph{Proceedings of the 18th Conference of the European Chapter of the Association for Computational Linguistics (Volume 1: Long Papers)}, pages 1689--1700, St. Julian{'}s, Malta. Association for Computational Linguistics.

\bibitem[{Kwon et~al.(2023)Kwon, Li, Zhuang, Sheng, Zheng, Yu, Gonzalez, Zhang, and Stoica}]{kwon2023efficient}
Woosuk Kwon, Zhuohan Li, Siyuan Zhuang, Ying Sheng, Lianmin Zheng, Cody~Hao Yu, Joseph~E. Gonzalez, Hao Zhang, and Ion Stoica. 2023.
\newblock Efficient memory management for large language model serving with pagedattention.
\newblock In \emph{Proceedings of the ACM SIGOPS 29th Symposium on Operating Systems Principles}.

\bibitem[{Madaan et~al.(2024)Madaan, Tandon, Gupta, Hallinan, Gao, Wiegreffe, Alon, Dziri, Prabhumoye, Yang et~al.}]{madaan2024self}
Aman Madaan, Niket Tandon, Prakhar Gupta, Skyler Hallinan, Luyu Gao, Sarah Wiegreffe, Uri Alon, Nouha Dziri, Shrimai Prabhumoye, Yiming Yang, et~al. 2024.
\newblock Self-refine: Iterative refinement with self-feedback.
\newblock \emph{Advances in Neural Information Processing Systems}, 36.

\bibitem[{OpenAI et~al.(2024)OpenAI, Achiam, Adler, Agarwal, Ahmad, Akkaya, Aleman, Almeida, Altenschmidt, Altman, Anadkat, Avila, Babuschkin, Balaji, Balcom, Baltescu, Bao, Bavarian, Belgum, Bello, Berdine, Bernadett-Shapiro, Berner, Bogdonoff, Boiko, Boyd, Brakman, Brockman, Brooks, Brundage, Button, Cai, Campbell, Cann, Carey, Carlson, Carmichael, Chan, Chang, Chantzis, Chen, Chen, Chen, Chen, Chen, Chess, Cho, Chu, Chung, Cummings, Currier, Dai, Decareaux, Degry, Deutsch, Deville, Dhar, Dohan, Dowling, Dunning, Ecoffet, Eleti, Eloundou, Farhi, Fedus, Felix, Fishman, Forte, Fulford, Gao, Georges, Gibson, Goel, Gogineni, Goh, Gontijo-Lopes, Gordon, Grafstein, Gray, Greene, Gross, Gu, Guo, Hallacy, Han, Harris, He, Heaton, Heidecke, Hesse, Hickey, Hickey, Hoeschele, Houghton, Hsu, Hu, Hu, Huizinga, Jain, Jain, Jang, Jiang, Jiang, Jin, Jin, Jomoto, Jonn, Jun, Kaftan, Łukasz Kaiser, Kamali, Kanitscheider, Keskar, Khan, Kilpatrick, Kim, Kim, Kim, Kirchner, Kiros, Knight, Kokotajlo, Łukasz Kondraciuk,
  Kondrich, Konstantinidis, Kosic, Krueger, Kuo, Lampe, Lan, Lee, Leike, Leung, Levy, Li, Lim, Lin, Lin, Litwin, Lopez, Lowe, Lue, Makanju, Malfacini, Manning, Markov, Markovski, Martin, Mayer, Mayne, McGrew, McKinney, McLeavey, McMillan, McNeil, Medina, Mehta, Menick, Metz, Mishchenko, Mishkin, Monaco, Morikawa, Mossing, Mu, Murati, Murk, Mély, Nair, Nakano, Nayak, Neelakantan, Ngo, Noh, Ouyang, O'Keefe, Pachocki, Paino, Palermo, Pantuliano, Parascandolo, Parish, Parparita, Passos, Pavlov, Peng, Perelman, de~Avila Belbute~Peres, Petrov, de~Oliveira~Pinto, Michael, Pokorny, Pokrass, Pong, Powell, Power, Power, Proehl, Puri, Radford, Rae, Ramesh, Raymond, Real, Rimbach, Ross, Rotsted, Roussez, Ryder, Saltarelli, Sanders, Santurkar, Sastry, Schmidt, Schnurr, Schulman, Selsam, Sheppard, Sherbakov, Shieh, Shoker, Shyam, Sidor, Sigler, Simens, Sitkin, Slama, Sohl, Sokolowsky, Song, Staudacher, Such, Summers, Sutskever, Tang, Tezak, Thompson, Tillet, Tootoonchian, Tseng, Tuggle, Turley, Tworek, Uribe, Vallone,
  Vijayvergiya, Voss, Wainwright, Wang, Wang, Wang, Ward, Wei, Weinmann, Welihinda, Welinder, Weng, Weng, Wiethoff, Willner, Winter, Wolrich, Wong, Workman, Wu, Wu, Wu, Xiao, Xu, Yoo, Yu, Yuan, Zaremba, Zellers, Zhang, Zhang, Zhao, Zheng, Zhuang, Zhuk, and Zoph}]{openai2024gpt4technicalreport}
OpenAI, Josh Achiam, Steven Adler, Sandhini Agarwal, Lama Ahmad, Ilge Akkaya, Florencia~Leoni Aleman, Diogo Almeida, Janko Altenschmidt, Sam Altman, Shyamal Anadkat, Red Avila, Igor Babuschkin, Suchir Balaji, Valerie Balcom, Paul Baltescu, Haiming Bao, Mohammad Bavarian, Jeff Belgum, Irwan Bello, Jake Berdine, Gabriel Bernadett-Shapiro, Christopher Berner, Lenny Bogdonoff, Oleg Boiko, Madelaine Boyd, Anna-Luisa Brakman, Greg Brockman, Tim Brooks, Miles Brundage, Kevin Button, Trevor Cai, Rosie Campbell, Andrew Cann, Brittany Carey, Chelsea Carlson, Rory Carmichael, Brooke Chan, Che Chang, Fotis Chantzis, Derek Chen, Sully Chen, Ruby Chen, Jason Chen, Mark Chen, Ben Chess, Chester Cho, Casey Chu, Hyung~Won Chung, Dave Cummings, Jeremiah Currier, Yunxing Dai, Cory Decareaux, Thomas Degry, Noah Deutsch, Damien Deville, Arka Dhar, David Dohan, Steve Dowling, Sheila Dunning, Adrien Ecoffet, Atty Eleti, Tyna Eloundou, David Farhi, Liam Fedus, Niko Felix, Simón~Posada Fishman, Juston Forte, Isabella Fulford, Leo
  Gao, Elie Georges, Christian Gibson, Vik Goel, Tarun Gogineni, Gabriel Goh, Rapha Gontijo-Lopes, Jonathan Gordon, Morgan Grafstein, Scott Gray, Ryan Greene, Joshua Gross, Shixiang~Shane Gu, Yufei Guo, Chris Hallacy, Jesse Han, Jeff Harris, Yuchen He, Mike Heaton, Johannes Heidecke, Chris Hesse, Alan Hickey, Wade Hickey, Peter Hoeschele, Brandon Houghton, Kenny Hsu, Shengli Hu, Xin Hu, Joost Huizinga, Shantanu Jain, Shawn Jain, Joanne Jang, Angela Jiang, Roger Jiang, Haozhun Jin, Denny Jin, Shino Jomoto, Billie Jonn, Heewoo Jun, Tomer Kaftan, Łukasz Kaiser, Ali Kamali, Ingmar Kanitscheider, Nitish~Shirish Keskar, Tabarak Khan, Logan Kilpatrick, Jong~Wook Kim, Christina Kim, Yongjik Kim, Jan~Hendrik Kirchner, Jamie Kiros, Matt Knight, Daniel Kokotajlo, Łukasz Kondraciuk, Andrew Kondrich, Aris Konstantinidis, Kyle Kosic, Gretchen Krueger, Vishal Kuo, Michael Lampe, Ikai Lan, Teddy Lee, Jan Leike, Jade Leung, Daniel Levy, Chak~Ming Li, Rachel Lim, Molly Lin, Stephanie Lin, Mateusz Litwin, Theresa Lopez, Ryan
  Lowe, Patricia Lue, Anna Makanju, Kim Malfacini, Sam Manning, Todor Markov, Yaniv Markovski, Bianca Martin, Katie Mayer, Andrew Mayne, Bob McGrew, Scott~Mayer McKinney, Christine McLeavey, Paul McMillan, Jake McNeil, David Medina, Aalok Mehta, Jacob Menick, Luke Metz, Andrey Mishchenko, Pamela Mishkin, Vinnie Monaco, Evan Morikawa, Daniel Mossing, Tong Mu, Mira Murati, Oleg Murk, David Mély, Ashvin Nair, Reiichiro Nakano, Rajeev Nayak, Arvind Neelakantan, Richard Ngo, Hyeonwoo Noh, Long Ouyang, Cullen O'Keefe, Jakub Pachocki, Alex Paino, Joe Palermo, Ashley Pantuliano, Giambattista Parascandolo, Joel Parish, Emy Parparita, Alex Passos, Mikhail Pavlov, Andrew Peng, Adam Perelman, Filipe de~Avila Belbute~Peres, Michael Petrov, Henrique~Ponde de~Oliveira~Pinto, Michael, Pokorny, Michelle Pokrass, Vitchyr~H. Pong, Tolly Powell, Alethea Power, Boris Power, Elizabeth Proehl, Raul Puri, Alec Radford, Jack Rae, Aditya Ramesh, Cameron Raymond, Francis Real, Kendra Rimbach, Carl Ross, Bob Rotsted, Henri Roussez,
  Nick Ryder, Mario Saltarelli, Ted Sanders, Shibani Santurkar, Girish Sastry, Heather Schmidt, David Schnurr, John Schulman, Daniel Selsam, Kyla Sheppard, Toki Sherbakov, Jessica Shieh, Sarah Shoker, Pranav Shyam, Szymon Sidor, Eric Sigler, Maddie Simens, Jordan Sitkin, Katarina Slama, Ian Sohl, Benjamin Sokolowsky, Yang Song, Natalie Staudacher, Felipe~Petroski Such, Natalie Summers, Ilya Sutskever, Jie Tang, Nikolas Tezak, Madeleine~B. Thompson, Phil Tillet, Amin Tootoonchian, Elizabeth Tseng, Preston Tuggle, Nick Turley, Jerry Tworek, Juan Felipe~Cerón Uribe, Andrea Vallone, Arun Vijayvergiya, Chelsea Voss, Carroll Wainwright, Justin~Jay Wang, Alvin Wang, Ben Wang, Jonathan Ward, Jason Wei, CJ~Weinmann, Akila Welihinda, Peter Welinder, Jiayi Weng, Lilian Weng, Matt Wiethoff, Dave Willner, Clemens Winter, Samuel Wolrich, Hannah Wong, Lauren Workman, Sherwin Wu, Jeff Wu, Michael Wu, Kai Xiao, Tao Xu, Sarah Yoo, Kevin Yu, Qiming Yuan, Wojciech Zaremba, Rowan Zellers, Chong Zhang, Marvin Zhang, Shengjia
  Zhao, Tianhao Zheng, Juntang Zhuang, William Zhuk, and Barret Zoph. 2024.
\newblock \href {http://arxiv.org/abs/2303.08774} {Gpt-4 technical report}.

\bibitem[{Qiu et~al.(2024)Qiu, Zhao, Ziser, Korhonen, Ponti, and Cohen}]{qiu-etal-2024-large}
Yifu Qiu, Zheng Zhao, Yftah Ziser, Anna Korhonen, Edoardo Ponti, and Shay Cohen. 2024.
\newblock \href {https://doi.org/10.18653/v1/2024.naacl-long.391} {Are large language model temporally grounded?}
\newblock In \emph{Proceedings of the 2024 Conference of the North American Chapter of the Association for Computational Linguistics: Human Language Technologies (Volume 1: Long Papers)}, pages 7064--7083, Mexico City, Mexico. Association for Computational Linguistics.

\bibitem[{Sakaguchi et~al.(2021)Sakaguchi, Bhagavatula, Le~Bras, Tandon, Clark, and Choi}]{sakaguchi-etal-2021-proscript-partially}
Keisuke Sakaguchi, Chandra Bhagavatula, Ronan Le~Bras, Niket Tandon, Peter Clark, and Yejin Choi. 2021.
\newblock \href {https://doi.org/10.18653/v1/2021.findings-emnlp.184} {pro{S}cript: Partially ordered scripts generation}.
\newblock In \emph{Findings of the Association for Computational Linguistics: EMNLP 2021}, pages 2138--2149, Punta Cana, Dominican Republic. Association for Computational Linguistics.

\bibitem[{Shinn et~al.(2024)Shinn, Cassano, Gopinath, Narasimhan, and Yao}]{shinn2024reflexion}
Noah Shinn, Federico Cassano, Ashwin Gopinath, Karthik Narasimhan, and Shunyu Yao. 2024.
\newblock Reflexion: Language agents with verbal reinforcement learning.
\newblock \emph{Advances in Neural Information Processing Systems}, 36.

\bibitem[{Sprague et~al.(2024)Sprague, Yin, Rodriguez, Jiang, Wadhwa, Singhal, Zhao, Ye, Mahowald, and Durrett}]{sprague2024cotmainlymath}
Zayne Sprague, Fangcong Yin, Juan~Diego Rodriguez, Dongwei Jiang, Manya Wadhwa, Prasann Singhal, Xinyu Zhao, Xi~Ye, Kyle Mahowald, and Greg Durrett. 2024.
\newblock To cot or not to cot? chain-of-thought helps mainly on math and symbolic reasoning.
\newblock \emph{arXiv preprint arXiv:2409.12183}.

\bibitem[{Storaï and Hwang(2024)}]{storaï2024harphesitationawarereframingtransformer}
Romain Storaï and Seung-won Hwang. 2024.
\newblock \href {http://arxiv.org/abs/2412.07282} {Harp: Hesitation-aware reframing in transformer inference pass}.

\bibitem[{Su et~al.(2024)Su, Zhang, Zhu, Qu, Li, Zhang, and Cheng}]{su2024timo}
Zhaochen Su, Jun Zhang, Tong Zhu, Xiaoye Qu, Juntao Li, Min Zhang, and Yu~Cheng. 2024.
\newblock Timo: Towards better temporal reasoning for language models.
\newblock \emph{arXiv preprint arXiv:2406.14192}.

\bibitem[{Thukral et~al.(2021)Thukral, Kukreja, and Kavouras}]{thukral2021probing}
Shivin Thukral, Kunal Kukreja, and Christian Kavouras. 2021.
\newblock Probing language models for understanding of temporal expressions.
\newblock In \emph{Proceedings of the Fourth BlackboxNLP Workshop on Analyzing and Interpreting Neural Networks for NLP}, pages 396--406.

\bibitem[{Wang et~al.(2023)Wang, Wei, Schuurmans, Le, Chi, Narang, Chowdhery, and Zhou}]{wang2023self}
Xuezhi Wang, Jason Wei, Dale Schuurmans, Quoc~V Le, Ed~H Chi, Sharan Narang, Aakanksha Chowdhery, and Denny Zhou. 2023.
\newblock Self-consistency improves chain of thought reasoning in language models.
\newblock In \emph{The Eleventh International Conference on Learning Representations}.

\bibitem[{Wang and Culotta(2020)}]{wang2020identifying}
Zhao Wang and Aron Culotta. 2020.
\newblock Identifying spurious correlations for robust text classification.
\newblock In \emph{Findings of the Association for Computational Linguistics: EMNLP 2020}, pages 3431--3440.

\bibitem[{Wei et~al.(2022)Wei, Wang, Schuurmans, Bosma, Xia, Chi, Le, Zhou et~al.}]{wei2022chain}
Jason Wei, Xuezhi Wang, Dale Schuurmans, Maarten Bosma, Fei Xia, Ed~Chi, Quoc~V Le, Denny Zhou, et~al. 2022.
\newblock Chain-of-thought prompting elicits reasoning in large language models.
\newblock \emph{Advances in neural information processing systems}, 35:24824--24837.

\bibitem[{Wolf et~al.(2020)Wolf, Debut, Sanh, Chaumond, Delangue, Moi, Cistac, Rault, Louf, Funtowicz, Davison, Shleifer, von Platen, Ma, Jernite, Plu, Xu, Le~Scao, Gugger, Drame, Lhoest, and Rush}]{wolf-etal-2020-transformers}
Thomas Wolf, Lysandre Debut, Victor Sanh, Julien Chaumond, Clement Delangue, Anthony Moi, Pierric Cistac, Tim Rault, Remi Louf, Morgan Funtowicz, Joe Davison, Sam Shleifer, Patrick von Platen, Clara Ma, Yacine Jernite, Julien Plu, Canwen Xu, Teven Le~Scao, Sylvain Gugger, Mariama Drame, Quentin Lhoest, and Alexander Rush. 2020.
\newblock \href {https://doi.org/10.18653/v1/2020.emnlp-demos.6} {Transformers: State-of-the-art natural language processing}.
\newblock In \emph{Proceedings of the 2020 Conference on Empirical Methods in Natural Language Processing: System Demonstrations}, pages 38--45, Online. Association for Computational Linguistics.

\bibitem[{Yao(2024)}]{yao2024largelanguagemodelscontrastive}
Liang Yao. 2024.
\newblock \href {http://arxiv.org/abs/2403.08211} {Large language models are contrastive reasoners}.

\bibitem[{Zhang et~al.(2024)Zhang, Beauchamp, and Wang}]{zhang-etal-2024-narrative}
Xinliang~Frederick Zhang, Nicholas Beauchamp, and Lu~Wang. 2024.
\newblock \href {https://doi.org/10.18653/v1/2024.findings-emnlp.963} {Narrative-of-thought: Improving temporal reasoning of large language models via recounted narratives}.
\newblock In \emph{Findings of the Association for Computational Linguistics: EMNLP 2024}, pages 16507--16530, Miami, Florida, USA. Association for Computational Linguistics.

\bibitem[{Zhou et~al.(2019)Zhou, Khashabi, Ning, and Roth}]{zhou2019going}
Ben Zhou, Daniel Khashabi, Qiang Ning, and Dan Roth. 2019.
\newblock “going on a vacation” takes longer than “going for a walk”: A study of temporal commonsense understanding.
\newblock In \emph{Proceedings of the 2019 Conference on Empirical Methods in Natural Language Processing and the 9th International Joint Conference on Natural Language Processing (EMNLP-IJCNLP)}, pages 3363--3369.

\bibitem[{Zhou et~al.(2021)Zhou, Richardson, Ning, Khot, Sabharwal, and Roth}]{zhou2021temporal}
Ben Zhou, Kyle Richardson, Qiang Ning, Tushar Khot, Ashish Sabharwal, and Dan Roth. 2021.
\newblock Temporal reasoning on implicit events from distant supervision.
\newblock In \emph{Proceedings of the 2021 Conference of the North American Chapter of the Association for Computational Linguistics: Human Language Technologies}, pages 1361--1371.

\bibitem[{Zhu et~al.(2023)Zhu, Yang, Chen, Li, Lou, and Yang}]{zhu2023question}
Xinyu Zhu, Cheng Yang, Bei Chen, Siheng Li, Jian-Guang Lou, and Yujiu Yang. 2023.
\newblock Question answering as programming for solving time-sensitive questions.
\newblock In \emph{Proceedings of the 2023 Conference on Empirical Methods in Natural Language Processing}, pages 12775--12790.

\end{thebibliography}
\bibliographystyle{acl_natbib}

\appendix

\newpage
\section*{\centering Appendices}

\section{Data Summary}
\label{appndx:data_summary}
Table~\ref{tab:data_summary} summarizes the dataset statistics used in this study. The numbers of official test samples are reported. 
Due to the budget, we evaluated Llama-3-8B on the full test set, GPT-4o-mini and Llama-3-70B on a random sample of up to 2,000 test set instances, and GPT-4o on 1,000 test set instances.

\begin{figure}[h]
    \centering
    \includegraphics[width=1\linewidth]{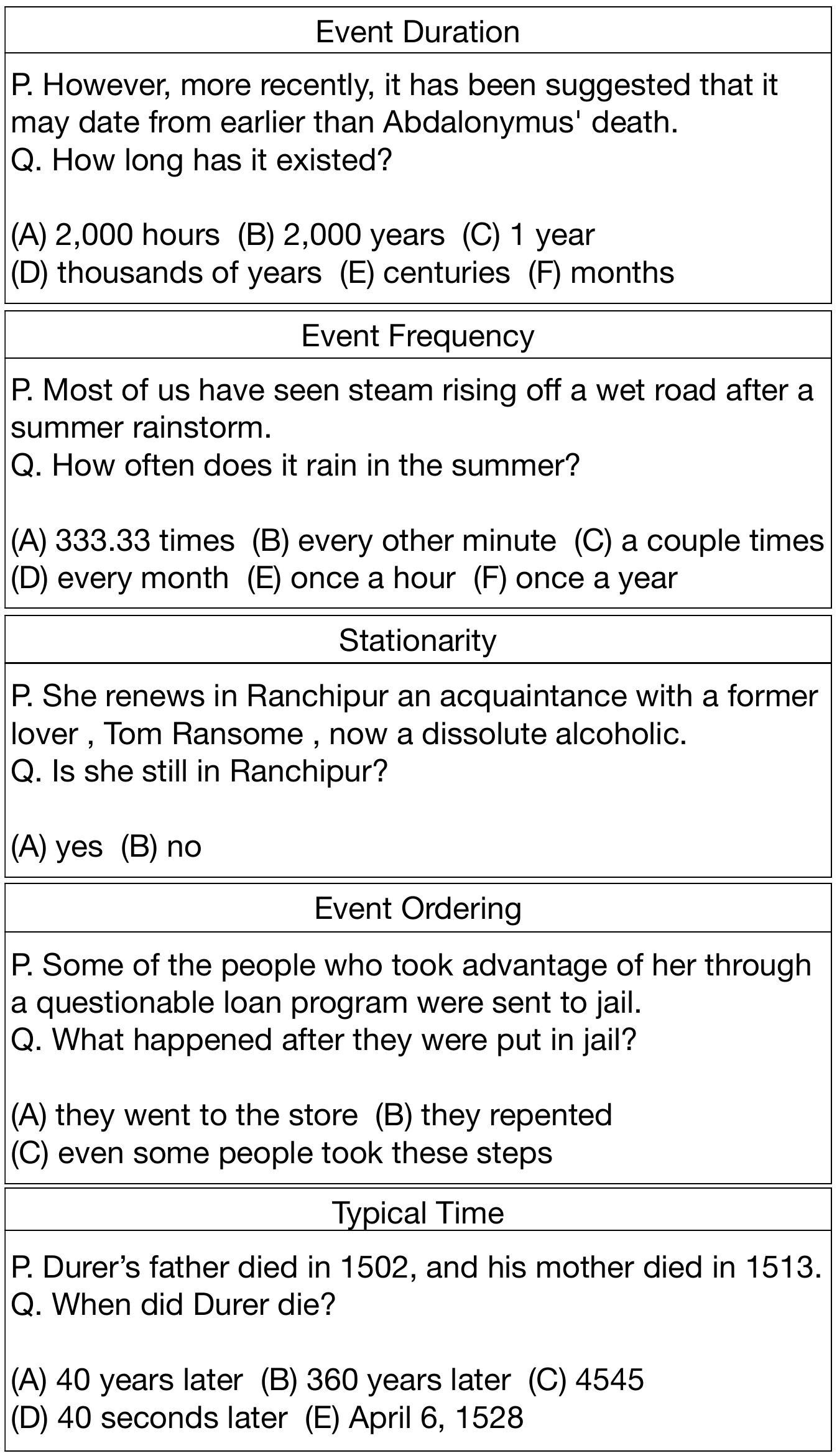}
    \caption{Examples of MCTACO Question Types. MCTACO covers various temporal aspects including event duration, frequency, stationarity, ordering, and typical time.}
    \label{fig:mctaco_examples}
\end{figure}

Additionally, the number of temporal relations considered in each dataset is included in Table~\ref{tab:data_summary}. TempEvalQA-Bi and TRACIE focus mainly on the before-after relation. MCTACO includes diverse temporal relations, and the number of annotated candidates is reported.
The questions in MCTACO are categorized into 5 question types, and examples for each type are provided in Figure~\ref{fig:mctaco_examples}.

\begin{table}[h]
\centering
\scalebox{0.95}{
\begin{tabular}{c|cc}
\toprule
                             & \#Test      & \#Temporal relations   \\ \midrule
TempEvalQA-Bi                & 448      &  2           \\ 
 TRACIE                      & 4248     &  2          \\
 MCTACO                      & 9442     & 1-19 \\ \bottomrule
\end{tabular}
}
\caption{Dataset Statistics.
For TempEvalQA-Bi, the numbers represent the total number of questions. For TRACIE, the numbers refer to the number of story-hypothesis pairs. For MCTACO, the numbers reflect question-and-answer candidate pairs.}
\label{tab:data_summary}
\end{table}

\section{Details of Evaluation Settings}
\label{appndx:hyperparam}

This section outlines the detailed evaluation settings, including hyperparameters, resources, efficiency, and parsing methods.
We use greedy decoding for \textbf{SP}, \textbf{CoT}, and \textbf{CCP}. For \textbf{Consistency}, \textbf{Reflect}, and \textbf{Debate}, we adopt the approach from~\citet{wang2023self}, employing top-k sampling with 
$k=40$ and a temperature of 0.5 for the LLaMA model. For GPT-based models, we set the temperature to 0.7.
\textbf{Consistency} samples 40 outputs from the decoder. \textbf{Reflect} refines the output iteratively for two iterations, including the initial output. In \textbf{Debate}, three agents engage in a debate over two rounds\cite{du2024improving}.
The implementations of the latter two baselines (\textbf{Reflect}, \textbf{Debate}) are based on the GitHub repository~\footnote{\url{https://github.com/composable-models/llm_multiagent_debate}} from~\citet{du2024improving}. Single-run performances are reported.

We note that our method prompts 3 times: for counterfactual question generation, counterfactual answer generation, and original question's answer generation, whose efficiency is compatible with or even more efficient than the three baselines. We also note that the Consistency baseline of Llama-3-70B cannot be reported due to its computation inefficiency.

For resources, we used the Transformers library~\cite{wolf-etal-2020-transformers} and vLLM~\cite{kwon2023efficient} with 4 RTX A6000 GPUs for Llama-3 models. We used Openai API~\footnote{\url{platform.openai.com}} for GPT models.
For output parsing, the models generate the final answer after the phrase \texttt{``Final answer:''}. Counterfactual exemplars are generated by modifying each dataset's questions, hypotheses, and candidate answers.

\begin{table}[h]
\centering
\scalebox{1.0}{
\begin{tabular}{c|c|c|ccc}
\toprule
Models & Methods   & \multicolumn{2}{c}{MCTACO}    \\ \midrule
       &           &   EM         & F1              \\ \midrule
GPT-4o  & MCQA-CoT  & 51.7        & 67.7        \\ 
-mini   &  \ours   & \textbf{58.9} & \textbf{78.6}  \\ \midrule
GPT-4o       & MCQA-CoT  & 65.6        & 73.3           \\ 
       &  \ours    & \textbf{66.2} & \textbf{80.2}    \\
 \bottomrule
\end{tabular}
}
\caption{Performance comparison on MCTACO with multiple-choice question answering setting.}
\label{tab:mcqa}
\end{table}



\section{Further Analysis}

\subsection{Task generalizability}
To demonstrate that our solution extends beyond binary question answering to multiple-choice question answering (MCQA), we evaluated the performance of GPT models using the original MCTACO evaluation setting~\cite{zhou2019going}. While our primary evaluation decomposed the multiple-choice format into binary questions to measure inconsistency, it can be reconstructed for multiple-choice evaluation. We additionally introduced a baseline for MCQA (MCQA-CoT) that provides the context, question, and all candidate answers, generating one or more correct answers step-by-step. 
The results in Table~\ref{tab:mcqa} indicate that our method (CCP) outperforms the MCQA-CoT baseline on multiple-choice tasks, demonstrating its effectiveness in the MCQA setting.


\subsection{Generated vs Retrieved Questions}

\begin{figure}[h]
    \centering
    \includegraphics[width=1\linewidth]{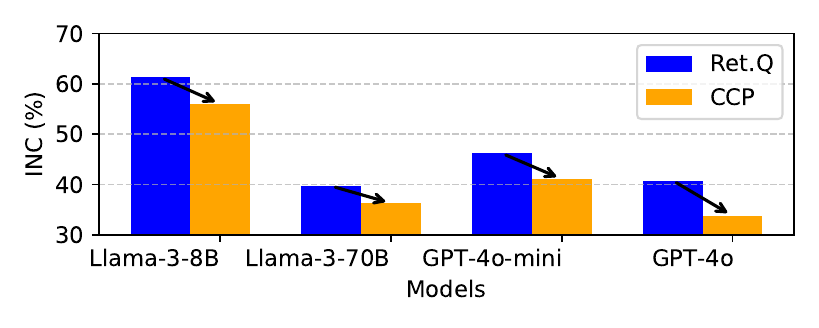}
    \vspace{-8mm}
    \caption{Comparison between counterfactual example collection methods on MCTACO with different models.}
    \label{fig:mctaco_retq_genq}
    \vspace{-3mm}
\end{figure}

We tested whether our claim in Figure~\ref{fig:mctaco_each_type} can be generalized to other models. Figure~\ref{fig:mctaco_retq_genq} consistently confirms that creating counterfactual questions by generation handles diverse temporal relations better than retrieving questions across different models.

\begin{figure}[h]
    \centering
    \includegraphics[width=1\linewidth]{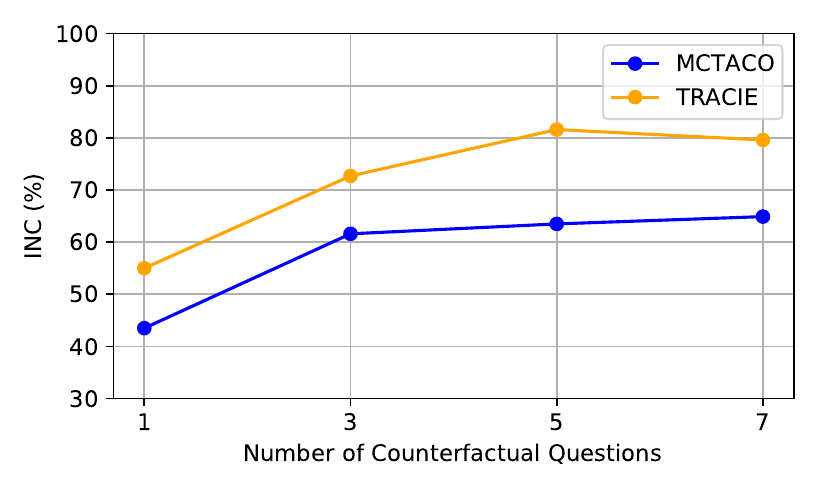}
    \vspace{-8mm}
    \caption{Inconsistency changes with the different number of counterfactual questions. The Llama-3-8B model is used.}
    \label{fig:num_cq}
    \vspace{-3mm}
\end{figure}

\subsection{Number of Counterfactual Questions}
We analyze the impact of varying the number of counterfactual questions on performance by testing with 1, 3, 5, and 7 questions. As shown in Figure~\ref{fig:num_cq}, performance is highest with a single counterfactual question, with degradation observed as the number increases. The result aligns with findings from prior studies, where incorporating excessive counterfactual or contrastive information in prompts often results in diminished performance~\cite{yao2024largelanguagemodelscontrastive,fang2024counterfactualdebatingpresetstances,storaï2024harphesitationawarereframingtransformer}. For instance, multiple information degrade the performance in arithmetic and symbolic reasoning~\cite{yao2024largelanguagemodelscontrastive}, multi-hop and commonsense question answering~\cite{fang2024counterfactualdebatingpresetstances}, and speculative decoding~\cite{storaï2024harphesitationawarereframingtransformer}. These results indicate that such a phenomenon is not specific to our method but highlights a broader limitation in LLMs' ability to process multiple counterfactual information effectively.

\begin{table}[ht]
\centering
\scalebox{0.85}{
\begin{tabular}{cc|cc|cc}
\toprule
            &             & \multicolumn{2}{c}{TimeQA} & \multicolumn{2}{c}{TimexNLI}        \\
\midrule
            &             & ACC         & F1           & ACC & F1  \\
\midrule
Llama & 3 shot       & \textbf{34.3} & 40.8          & 68.0                    & 65.3            \\
 3-8B      & CoT 3 shot   & 32.3          & 38.4          & \textbf{74.0}                    & \textbf{73.3}          \\
            & \ours 3 shot   & 34            & \textbf{41.5} &            67.3       &   62.2 \\
\midrule
GPT-4o & 3 shot       & 40                                & 52.36                             & 86.4                    & 85.3             \\
 -mini     & CoT 3 shot   & \textbf{43.3}                     & \textbf{56.75}                    & \textbf{90.4}           & \textbf{90.3}        \\
            & \ours 3 shot  & 41                                & 53.59                             & 90.3                    & 90.0          \\ 
\bottomrule
\end{tabular}
}
\caption{Performance comparison on TimeQA and TimexNLI.}
\label{tab:bad}
\end{table}

\section{Details of Limitations}

\subsection{Tasks with temporal indicators}
\label{appndx:detail_limitation_temporal_indicator}
Table~\ref{tab:bad} shows the experimental results for the tasks requiring the understanding of temporal indicators. We evaluated our method on TimeQA~\cite{chen2dataset}, the event-time ordering task, and TimexNLI-T1~\cite{thukral2021probing}, the time-time ordering task, where \ours showed limited performance gains.

\subsection{Prompt Sensitivity}
\label{appndx:detail_limitation_prompt_sensitivity}
\begin{table*}[h]
    \centering
    \renewcommand{\arraystretch}{1.2}
    \setlength{\tabcolsep}{5pt}
    \begin{tabular}{l |l |c c c c c c c c c}
        \toprule
        \multirow{2}{*}{Models} & \multirow{2}{*}{Prompts} & \multicolumn{3}{c}{TempEvalQA-Bi} & \multicolumn{3}{c}{TRACIE} & \multicolumn{3}{c}{MCTACO} \\
        \cmidrule(lr){3-5} \cmidrule(lr){6-8} \cmidrule(lr){9-11}
        & & ACC & F1 & INC & ACC & F1 & INC & ACC & F1 & INC \\
        \midrule
        Llama-3-8B & CoT & 69.6 & 70.6 & 50.0 & 63.0 & 64.9 & 56.0 & 77.6 & 69.8 & 63.4 \\
                   & CCP (e2e) & \underline{75.4} & \underline{73.6} & \underline{33.9} & \underline{66.8} & \underline{65.4} & \underline{46.7} & \textbf{83.1} & \textbf{81.0} & \textbf{53.7} \\
                   & CCP        & \textbf{75.9} & \textbf{75.2} & \textbf{32.7} & \textbf{68.8} & \textbf{70.4} & \textbf{39.8} & \underline{78.7} & \underline{81.7} & \underline{57.7} \\ \hline
        GPT-4o-mini & CoT & \underline{81.3} & 79.9 & 29.0 & 73.2 & 68.5 & 42.7 & 80.9 & 73.7 & 58.9 \\
                    & CCP (e2e) & 80.8 & \underline{80.2} & \underline{27.7} & \underline{79.0} & \underline{78.3} & \underline{28.0} & \underline{87.1} & \underline{84.4} & \underline{52.1} \\
                    & CCP      & \textbf{88.8} & \textbf{88.7} & \textbf{19.6} & \textbf{82.5} & \textbf{81.2} & \textbf{20.2} & \textbf{87.8} & \textbf{85.8} & \textbf{42.1} \\
        \bottomrule
    \end{tabular}
    \caption{Performance comparison with different prompts. The best and second-best performances are in boldface and underlined, respectively.}
    \label{tab:prompt_sensitivity}
\end{table*}
Our key decision in the prompt design was to separate the counterfactual question generation prompt (CCP), described in Appendix~\ref{appndx:prompt_examples_gen_counter_q}, from the prompt that answers counterfactual and original questions, detailed in Appendix~\ref{appndx:prompt_examples_gen_ans}. If we generate counterfactual questions and answers in an end-to-end manner using only the prompt in Appendix~\ref{appndx:prompt_examples_gen_ans} (CCP(e2e)), while it still outperforms CoT, the performance slightly decreases, as shown in Table~\ref{tab:prompt_sensitivity}.

\section{Usage of AI Assistants}
ChatGPT was employed to generate answers in the prompt examples.

\newpage

\section{Prompt Templates}
We list the prompts that we used.

\subsection{Prompt Templates for Generating Counterfactual Questions}
\label{appndx:prompt_examples_gen_counter_q}

To generate the counterfactual questions, we use the prompts provided below.
We control question aspects by tailoring ICL examples and prompting the model to follow few-shot examples with the prompt: (\texttt{``following previous examples''}).

\begin{myColorbox}[]{TempEvalQA-Bi}

\textbf{[System]}

You are an expert in temporal knowledge. Generate one related question to help determine the correctness of the original question, following the previous examples.

\textbf{[User]}

Question: Is Farkas sent into space on board the Soyuz before McBride on board the Orbiter Challenger?

\textbf{[Assistant]}

Is Farkas sent into space on board the Soyuz after McBride on board the Orbiter Challenger?

\textbf{[User]}

Question: Is McBride on board the Orbiter Challenger after Farkas was made a brigadier general?

\textbf{[Assistant]}

Is McBride on board the Orbiter Challenger before Farkas was made a brigadier general?

\textbf{[User]}

Question: Is Farkas was appointed military attache at the Hungarian embassy before he was made a brigadier?

\textbf{[Assistant]}

Is Farkas was appointed military attache at the Hungarian embassy after he was made a brigadier?
\end{myColorbox}

\begin{myColorbox}[]{TRACIE}

\textbf{[System]}

You are an expert in temporal knowledge. Generate one related hypothesis to help determine the plausibility of the original hypothesis, following previous examples.

\textbf{[User]}

Hypothesis: Chad looked for his baseball cap starts after he got off the ride.

\textbf{[Assistant]}

Chad looked for his baseball cap starts before he got off the ride.

\textbf{[User]}

Hypothesis: Tom's friend says he is not gay starts before One day he noticed someone else asking

\textbf{[Assistant]}

Tom's friend says he is not gay starts after One day he noticed someone else asking

\textbf{[User]}

Hypothesis: Liza read a book on economics in the library starts before she got to the library

\textbf{[Assistant]}

Liza read a book on economics in the library starts after she got to the library 

\end{myColorbox}

\begin{myColorbox}[]{MCTACO}

\textbf{[System]}

You are an expert in temporal commonsense knowledge. Generate related candidate answers that are either plausible or implausible that help determine the correctness of the original candidate answer following the previous examples.

\textbf{[User]}

Passage: The majority religion during the centuries of Ottoman rule, though a significant Christian minority remained. Question: What happened before Islam was the majority religion? Candidate answer: christianity was the majority religion.

\textbf{[Assistant]}

The Ottoman Empire had just begun /\ The spread of Byzantine influence in the region /\ The emergence of Islam as a minor religion 

\textbf{[User]}

Passage: It's hail crackled across the comm, and Tara spun to retake her seat at the helm. Question: How long was the storm? Candidate answer: 6 years.

\textbf{[Assistant]}

an hour /\ a week /\ a month

\textbf{[User]}

Passage: Passage: About 30\% of Ratners's profit already is derived from the U.S. Question: Is Ratners's profit derived from the U.S. today? Candidate answer: yes.

\textbf{[Assistant]}

no

\end{myColorbox}

\subsection{Prompt Templates for CCP and Baselines}
\label{appndx:prompt_examples_gen_ans}
We only show the SP, CoT, and CCP since Consistency is the method that leverages CoT multiple times, and the Reflect and Debate follow the implementation of~\citet{du2024improving} after CoT. 

Given the system and user message, the model generates an assistant's message. In our approach, the model is prompted twice: first to create an assistant's answer to the counterfactual question, and then to generate an assistant's answer to the original question. Though we use 3-shot examples, we provide 1 example for simplicity.

Prompt templates used to generate answers for the TempevalQA-Bi dataset are provided below.

\begin{myColorbox}[]{TempevalQA-Bi: SP}

\textbf{[System]}

You are an expert in temporal knowledge. Answer the question according to the article. Answer the question yes or no (Format: final answer: yes/no)

\textbf{[User]}

Article: Farkas, an air force captain, was sent into space on board the Soyuz 36 on May 26, 1980. He spent six days aboard the Salyut 6 spacecraft with three Soviet astronauts, Valery Kubasov, Leonid Popov and Valery Riumin. McBride, 54, of Lewisburg, West Virginia, was part of a seven-member crew aboard the Orbiter Challenger [...] 

Question: Is Farkas sent into space on board the Soyuz before McBride on board the Orbiter Challenger? 
\newline 
\textbf{[Assistant]}
\newline
Final answer: yes
\end{myColorbox}

\begin{myColorbox}[]{TempevalQA-Bi: CoT}

\textbf{[System]}

You are an expert in temporal knowledge. Answer the question according to the article. Let's think step by step to answer the question. The answer should be "yes" or "no" followed by "Final answer: ".\newline
\textbf{[User]}

Article: [...] \newline
Question: Is Farkas sent into space on board the Soyuz before McBride on board the Orbiter Challenger? \newline 
\textbf{[Assistant]}
\newline
Let's think step by step.
Farkas was sent into space on May 26, 1980 and McBride was on board the Orbiter Challenger in October 1984. May 26, 1980 is before October 1984. So the answer is: yes.

Final answer: yes

\end{myColorbox}

\begin{myColorbox}[]{TempevalQA-Bi: CCP}

\textbf{[System]}

You are an expert in temporal knowledge. Answer the question according to the article. To improve understanding, follow these steps: Answer the related question to better understand the timeline (Format: answer: [yes/no]) Using the answers to the related questions as the additional information, answer the original question yes or no (Format: final answer: [yes/no]). \newline 
\textbf{[User]}

Article: [...]\newline
Original question: Is Farkas sent into space on board the Soyuz before McBride on board the Orbiter Challenger?\newline
Related question: Is Farkas sent into space on board the Soyuz after McBride on board the Orbiter Challenger?\newline 
Answer the related question\newline
\textbf{[Assistant]} \textit{\textcolor{blue}{\# Prediction for the counterfactual question}}

Farkas was sent into space on board the Soyuz 36 on May 26, 1980. McBride was part of the crew on board the Orbiter Challenger in October 1984. 
Since 1980 is before 1984, no, Farkas was not sent into space after McBride.\newline
answer: no\newline 
\textbf{[User]}
\newline
Now, answer the original question. Question: Is Farkas sent into space on board the Soyuz before McBride on board the Orbiter Challenger?\newline
\textbf{[Assistant]} \textit{\textcolor{blue}{\# Prediction for the original question}}

Farkas was sent into space on May 26, 1980 and McBride was on board the Orbiter Challenger in October 1984. May 26, 1980 is before October 1984. So the answer is: yes. \newline
Final answer: yes

\end{myColorbox}

Prompt templates used to generate answers for
the TRACIE dataset are provided below.

\begin{myColorbox}[]{TRACIE: SP}

\textbf{[System]}

You are an expert in temporal knowledge. You can understand the implicit temporal relationships between events to make judgments. Your task is to determine the relationship between the story and hypothesis. A `positive' label means the hypothesis can be inferred from the story or logically sound. A `negative' label means the hypothesis cannot be inferred from the story. You must answer only with the labels `positive' or `negative'. (Format: final answer: positive/negative)

\textbf{[User]}

Story: Chad looked for his baseball cap starts after he got off the ride story: Chad had gone to an amusement park. He was riding on the roller coaster. \newline
Hypothesis: Chad looked for his baseball cap starts after he got off the ride.
\newline 
\textbf{[Assistant]}
\newline
Final answer: positive
\end{myColorbox}

\begin{myColorbox}[]{TRACIE: CoT}

\textbf{[System]}

You are an expert in temporal knowledge. You can understand the implicit temporal relationships between events to make judgments. Your task is to determine the relationship between the story and hypothesis. A `positive' label means the hypothesis can be inferred from the story or logically sound. A `negative' label means the hypothesis cannot be inferred from the story. You must give final answer only with the labels `positive' or `negative' after the ``Final answer:'' prompt.
Let's think step by step to evaluate the relationship.

\textbf{[User]}

Story: [...]	\newline
Hypothesis: Chad looked for his baseball cap starts after he got off the ride.

\textbf{[Assistant]}

Let's think step by step: \newline
The story states that Chad's baseball cap fell off during the roller coaster ride.\newline
It also mentions that Chad found the cap after he got off the ride.
To find something, one typically needs to look for it first.
Since Chad found the cap after getting off the ride, he must have started looking for it after getting off the ride.
The hypothesis directly aligns with this sequence of events.
\newline
Final answer: positive

\end{myColorbox}

\begin{myColorbox}[]{TRACIE: CCP}

\textbf{[System]}

You are an expert in temporal knowledge. You can understand the implicit temporal relationships between events to make judgments. Your task is to determine the relationship between the story and hypothesis. A `positive' label means the hypothesis can be inferred from the story or logically sound. A `negative' label means the hypothesis cannot be inferred from the story. You must give final answer only with the labels `positive' or `negative'. 
To improve understanding, follow these steps: Evaluate the related hypothesis to help clarify the story's timeline (Format: answer: [positive/negative]). Using answers to the related hypothesis as the additional information, evaluate the original hypothesis (Format: final answer: [positive/negative]).

\textbf{[User]}

Story: [...] \newline
Hypothesis: Chad looked for his baseball cap starts after he got off the ride. \newline
Related hypothesis: Chad looked for his baseball cap starts before he got off the ride. \newline
Answer the related hypothesis

\textbf{[Assistant]}  \textit{\textcolor{blue}{\# Prediction for the counterfactual question}}

The hypothesis conflicts with the story, as Chad finds the cap after the ride.
\newline
Answer: negative
\newline 
\textbf{[User]}
\newline
Now, evaluate the original hypothesis: Chad looked for his baseball cap starts after he got off the ride.

\textbf{[Assistant]}  \textit{\textcolor{blue}{\# Prediction for the original question}}

Chad rides the roller coaster while wearing a baseball cap, which falls off during the ride. After getting off the ride, Chad finds the cap. The hypothesis suggests that Chad started looking for the cap after getting off the ride, which aligns with the story's sequence of events.
\newline
Final answer: positive

\end{myColorbox}

Prompt templates used to generate answers for the MCTACO dataset are provided below.

\begin{myColorbox}[]{MCTACO: SP}

\textbf{[System]}

You are expert in temporal commonsense knowledge. Your goal is to determine if the candidate answer to a given question is plausible based on the information in the passage. Answer the related candidate answers then answer the candidate answer. Provide a `yes' or `no' response to indicate whether the original candidate answer is correct. (Format: final answer: yes/no)
If the passage does not provide explicit details, provide an educated guess based on what is common or plausible in similar situations

\textbf{[User]}

Passage: the majority religion during the centuries of Ottoman rule, though a significant Christian minority remained. 

Question: What happened before Islam was the majority religion? 

Candidate answer: the end of white-minority rule.
\newline 
\textbf{[Assistant]}
\newline
Final answer: no
\end{myColorbox}

\begin{myColorbox}[]{MCTACO: CoT}

\textbf{[System]}

You are an expert in temporal commonsense knowledge. Your goal is to determine if the candidate answer to a given question is plausible based on the information in the passage. Let's think step by step and provide a final `yes' or `no' response to indicate whether the original candidate answer is correct. If the passage does not provide explicit details, provide an educated guess based on what is common or plausible in similar situations. The final answer should be in the format: ``Final answer: yes'' or ``Final answer: no''.

\textbf{[User]}

Passage: the majority religion during the centuries of Ottoman rule, though a significant Christian minority remained. 

Question: What happened before Islam was the majority religion? 

Candidate answer: christianity was the majority religion

\textbf{[Assistant]}

Let's think step by step.
\newline
Step 1: Analyze the passage.
The passage states that during the centuries of Ottoman rule, Islam was the majority religion, while a significant Christian minority remained. This implies that before the Ottoman Empire (which was Islamic), there could have been a different majority religion.
\newline
Step 2: Historical context.
Before the rise of the Ottoman Empire, large parts of the Middle East, North Africa, and parts of Europe were under the rule of the Byzantine Empire, which was a Christian empire. Additionally, many regions that later came under Ottoman rule were predominantly Christian.
\newline
Step 3: Plausibility of the candidate answer.
Given that Christianity was the dominant religion in many regions before the Ottoman Empire and its Islamic rule, it is plausible that Christianity was the majority religion before Islam in those areas.
\newline
Final answer: yes

\end{myColorbox}

\begin{myColorbox}[]{MCTACO: CCP}

\textbf{[System]}

You are an expert in temporal commonsense knowledge. Your goal is to determine if the candidate answer to a given question is plausible based on the information in the passage. To improve understanding, follow these steps: generate related candidate answers that are either plausible or implausible (Format: answer: yes/no) to help in determining the correctness of the original candidate answer. Provide a 'yes' or 'no' response to indicate whether the original candidate answer is correct (Format: final answer: yes/no). If the passage does not provide explicit details, provide an educated guess based on what is common or plausible in similar situations

\textbf{[User]}

Passage: the majority religion during the centuries of Ottoman rule, though a significant Christian minority remained. 

Question: What happened before Islam was the majority religion? 

original candidate answer: christianity was the majority religion
\newline
Related candidate: the end of white-minority rule
\newline 
Can the candidate answer the given question? yes or no

\textbf{[Assistant]}  \textit{\textcolor{blue}{\# Prediction for the counterfactual question}}

This is incorrect because the passage discusses events during Ottoman rule, not before it.
\newline
answer: no

\textbf{[User]}

Now, can the candidate answer the given question? candidate answer: christianity was the majority religion

\textbf{[Assistant]}  \textit{\textcolor{blue}{\# Prediction for the original question}}

This is plausible because Islam would have started to emerge before becoming the majority.

Final answer: yes

\end{myColorbox}

\section{Scientific Artifacts}

We used existing scientific artifacts for research purposes, and the use of existing artifacts was consistent with their intended applications.

TempEvalQA-Bi: \href{https://github.com/yfqiu-nlp/temporal-llms?tab=MIT-1-ov-file#readme}{MIT license}

TRACIE: \href{https://www.apache.org/licenses/LICENSE-2.0}{Apache-2.0 license}

Llama-3: \href{https://www.llama.com/llama3/license/}{custom commercial license}

OpenAI API: \href{https://www.apache.org/licenses/LICENSE-2.0}{Apache-2.0 license}

\end{document}